\documentclass[final]{cvpr}

\usepackage{times}
\usepackage{epsfig}
\usepackage{graphicx}
\usepackage{amsmath}
\usepackage{amssymb}

\usepackage{amsmath,amsfonts,bm}

\def\eqref#1{equation~\ref{#1}}

\def\1{\bm{1}}

\DeclareMathAlphabet{\mathsfit}{\encodingdefault}{\sfdefault}{m}{sl}
\SetMathAlphabet{\mathsfit}{bold}{\encodingdefault}{\sfdefault}{bx}{n}

\usepackage{array}
\usepackage{microtype}
\usepackage{graphicx}
\usepackage{subfigure}
\usepackage{booktabs} 
\newcolumntype{P}[1]{>{\centering\arraybackslash}p{#1}}
\newcommand{\centered}[1]{\begin{tabular}{c} #1 \end{tabular}}

\usepackage[pagebackref=true,breaklinks=true,colorlinks,bookmarks=false]{}

\def\cvprPaperID{****} 
\def\confYear{CVPR 2021}

\begin{document}



\title{Neural 3D Scene Compression via Model Compression }

\author{Berivan Isik\\
Stanford University\\
{\tt\small berivan.isik@stanford.edu}
}

\maketitle

\begin{abstract}
Rendering 3D scenes requires access to arbitrary viewpoints from the scene. Storage of such a 3D scene can be done in two ways; (1) storing 2D images taken from the 3D scene that can reconstruct the scene back through interpolations, or (2) storing a representation of the 3D scene itself that already encodes views from all directions. So far, traditional 3D compression methods have focused on the first type of storage and compressed the original 2D images with image compression techniques. With this approach, the user first decodes the stored 2D images and then renders the 3D scene. However, this separated procedure is inefficient since a large amount of 2D images have to be stored. In this work, we take a different approach and compress a functional representation of 3D scenes. In particular, we introduce a method to compress 3D scenes by compressing the neural networks that represent the scenes as
neural radiance fields. Our method provides more efficient storage of 3D scenes since it does not store 2D images -- which are redundant when we render the scene from the neural functional representation.\footnote{Stanford CS 231A Final Project, 2021.}

\end{abstract}

\section{Introduction}
We live in the multimedia era; communication and storage of text, audio, image, and video have become a vital part of our daily lives. The frequent use of this almost unlimited amount of multimedia data is possible thanks to developments in compression algorithms over the last century \cite{bhaskaran1997image, lewis1992image, moffat1989word, pan1995tutorial, rabbani2002jpeg2000}. Unlike text, audio, image, and video; communication and storage of 3D data have not attracted much attention for many years due to the lack of efficient 3D compression methods. However, 3D data is becoming more popular with recent improvements in augmented, virtual, and mixed reality experiences. This makes it critical to develop 3D compression techniques to enable more efficient communication and storage of 3D data in resource-constrained environments. \\ 

Existing 3D compression approaches focus on compressing the original 2D images that are normally used for rendering \cite{golts2020image, 7340499,peng2005technologies, siddeq2014novel}. After decoding those images back, rendering can be performed as usual. Unfortunately, this is a sub-optimal approach since it requires compressing multiple 2D images for one 3D scene. For a more efficient 3D compression procedure, we need to find solutions beyond just adapting existing 2D image compression algorithms for 3D. In this work, we put some effort towards this goal.\\


\begin{figure}[!t]
        \centering
        \subfigure[Original NeRF model.]{\includegraphics[width=.40\textwidth]{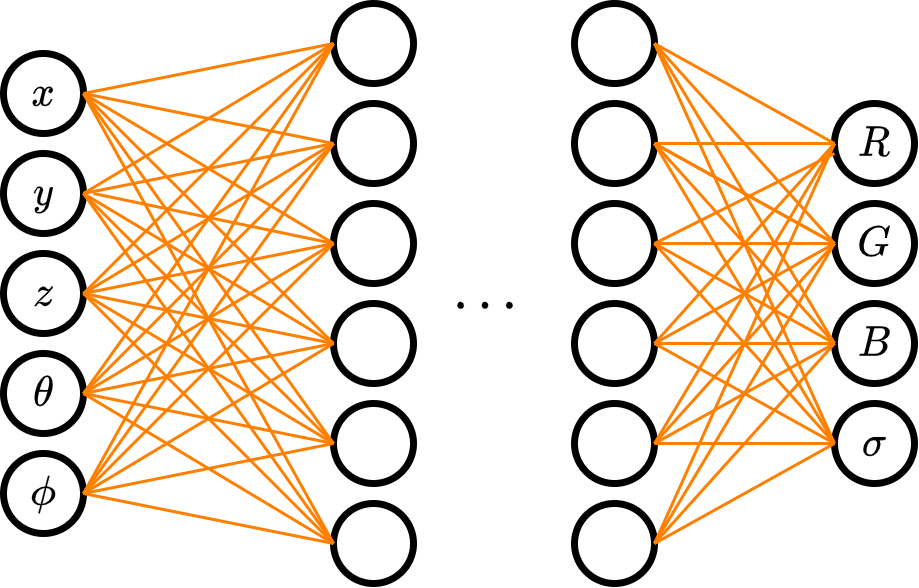}}
        \subfigure[Pruned NeRF model. ]{\includegraphics[width=.40\textwidth]{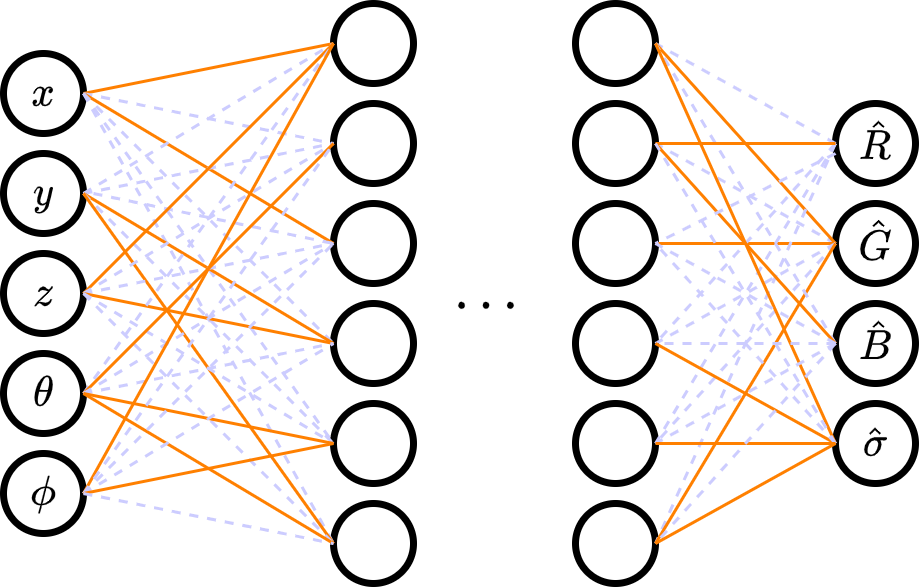}}
   \caption{Fully-connected NeRF model (a) before and (b) after pruning.}
\label{fig:pruning}
\end{figure}
Recent studies on deep learning for 3D scene representation have shown promising results in terms of perceptual quality of the rendered scene and efficiency of the representation \cite{mescheder2019occupancy,nerf,deepsdf, sitzmann2020implicit,nglod, tancik2020fourier}. Motivated by this, we primarily focus on the Neural Radiance Fields (NeRF) model \cite{nerf} which takes a 5D coordinate (the spatial location ($x$, $y$, $z$) and viewing direction ($\theta$, $\phi$)) as input and outputs a volume density and view-dependent emitted radiance at
that spatial location. We can think of the NeRF model as a functional representation of a 3D scene since it can generate all necessary information for rendering after training. Representing a 3D scene in a moderate-size model without needing 2D images has already been a 3D compression method. Our goal is to improve the compression ratio even further by compressing the NeRF model using neural network compression techniques \cite{cheng2017survey}.\\

To this end, we propose compressing 3D scene representations through model pruning as shown in Figure~\ref{fig:pruning}. Our method provides faster rendering, as well as more efficient storage and communication of 3D scenes.   
\section{Related Work}
Our work is most related to the fields of 3D compression, neural rendering, and neural network compression:
\paragraph{3D Compression.} 3D rendering is reconstructing 3D scenes in the form of point clouds, surface mesh models \cite{fuhrmann2014mve}, volumetric occupancy grids \cite{choy20163d}, or depth maps from 2D images. As these reconstructions occupy a huge memory, their storage and communication is costly. One workaround that many researchers have adopted to enable efficient transmission of 3D scenes is to transmit the 2D images that are required for rendering and leave the rendering process to the decoder. However, we need a large number of 2D images taken from the scene for a sufficiently good reconstruction. Thus, this separated approach --separating transmission and rendering-- requires 2D image compression. Prior work has mostly focused on image compression methods that perform well in terms of image quality metrics, but not necessarily quality metrics of the 3D rendering \cite{imagecomp}. More recently, optimization of image and video compression methods for 3D reconstruction has attracted interest by the research community \cite{astola2018wasp,golts2020image,jia2018light, zhao2018light}. In addition to the image compression approach, there has been a rising interest in the compression of 3D volumetric data itself, such as polygon surface meshes \cite{mesh4, mesh3, mesh2, mesh1}. In 2017, the Moving Picture Experts Group (MPEG) announced the first call for proposals for Point Cloud Compression \cite{pcc_mpeg}. Since then, research on the compression of point clouds has gained momentum  \cite{chou2019volumetric, kammerl2012real, pcc2}. 

\paragraph{Neural Rendering.} The recent success of deep learning in various applications such as image classification, language processing, and speech recognition has created an emerging field of research: deep learning for realistic rendering of 3D scenes. As a result, attempts on the implicit control of scene properties such as geometry, attribute, and camera parameters through neural functional representations have exploded \cite{tewari2020state}. In particular, researchers have been trying to encode a ``function'' from the 3D world such as spatial coordinates, or viewing direction, or both to image rendering quantities such as RGB colors, volume density, and signed distance function (SDF). This ``function'' is usually a neural network that is trained to improve 3D reconstruction quality. For instance, Occupancy Networks \cite{mescheder2019occupancy}, DeepSDF \cite{deepsdf}, and NGLOD \cite{nglod} encode an implicit surface representation of the scene; while Neural Volumes \cite{lombardi2019neural} and NeRF \cite{nerf} encode a coordinate-based 3D volume representation of the scene in a neural network. In this work, we focus on the NeRF model that encodes a function from the spatial coordinates $(x,y,z)$ and direction $(\theta, \phi)$ to RGB colors and volume density $\sigma$ at that location (see Figure~\ref{fig:pruning}(a)).

\paragraph{Neural Network Compression.} 
The interest in neural network compression research has increased sharply in the last decade \cite{han2015deep, he2018amc, isik2021neural, molchanov2017variational} as the growing size of neural networks becomes a significant bottleneck in their deployment on resource-constraint devices \cite{dean2012large, lecun2015deep}. Pruning is one of the neural network compression techniques that has been improved to the extent that today it is possible to prune more than $90 \%$ of the parameters of a trained model without a performance degradation \cite{gale2019state, han2015deep, han2015learning, isik2021successive}. An example of pruning a fraction of the parameters (connections) is shown in Figure~\ref{fig:pruning}. In this work, we introduce model pruning as a 3D compression method.


\section{Method}
\begin{figure*}[!t]
        \centering
        \subfigure[Colored view from the fern dataset. ]{\includegraphics[width=.30\textwidth]{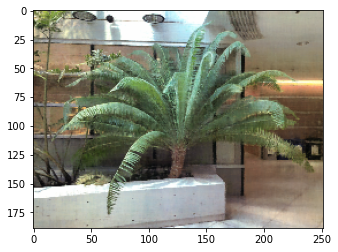}}
        \subfigure[Mesh model of the lego dataset. ]{\includegraphics[width=.30\textwidth]{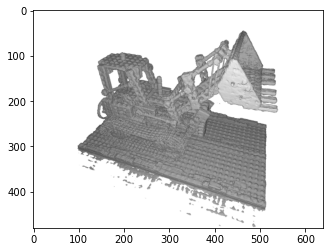}}
        \subfigure[Depth map of the lego dataset. ]{\includegraphics[width=.30\textwidth]{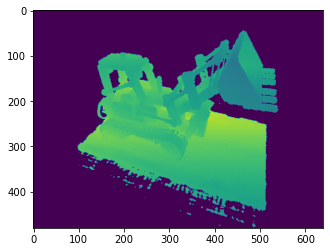}}
\caption{Rendered view, mesh model, and depth map of the fern and lego datasets using the original NeRF model (before pruning).}
\label{fig:originals}
\end{figure*}


We propose a novel 3D compression method by compressing the neural functional representation of the 3D scene. In particular, we consider the NeRF model \cite{nerf} as the implicit representation of the scene, although we could follow the same methodology for other neural functional representations as well. As the model compression technique, we choose magnitude pruning (removing the weights with smallest magnitudes) for several reasons: (1) NeRF model is a fully-connected network, and magnitude pruning has been shown to be successful, especially in fully-connected layers \cite{han2015deep}, (2) different from other neural network compression approaches, pruning speeds up the inference, which provides faster rendering in 3D reconstruction \cite{gale2019state}, and (3) pruning is an intuitive method with easy implementation. A visualization of our proposed scheme is given in Figure~\ref{fig:pruning}. We adopt global pruning approach as opposed to structured pruning, because global pruning has more flexibility to adjust the sparsity of different layers, which yields a better performance-compression trade-off. In the pruning literature, a common approach is to do iterative pruning with repeating cycles of pruning and retraining. In this work, we do one shot pruning followed by a short retraining process. However, our method can easily be extended to iterative pruning. \\

Note that pruning has been an effective neural network compression method for classification tasks. However, here, we apply pruning for a regression task to render realistic 3D scenes.  

\section{Experiments}

In our experiments, we tackled the following question: \\

\emph{Is it possible to compress 3D scene representations via model pruning and preserve the quality of reconstructions?} \\

\subsection{Datasets, Models, and Hyperparameters}
We used pre-trained NeRF models of the original paper trained on Lego and Fern datasets \footnote{\url{https://github.com/bmild/nerf}}. We pruned the models with various pruning ratios ($30\%$, $50\%$, $70\%$, and $90\%$). We present qualitative and quantitative comparisons with and without retraining after the pruning. The retraining duration is chosen as $500$ iterations and kept the same for both datasets in all pruning ratios. The hyperparameters during the retraining phase are the same as the hyperparameters of the corresponding iterations in the original NeRF paper. We applied global pruning, i.e., pruning threshold is the same for each layer. 
\subsection{Results}
\begin{figure}[!h]
        \centering
        \subfigure[PSNR for the fern dataset. ]{\includegraphics[width=.45\textwidth]{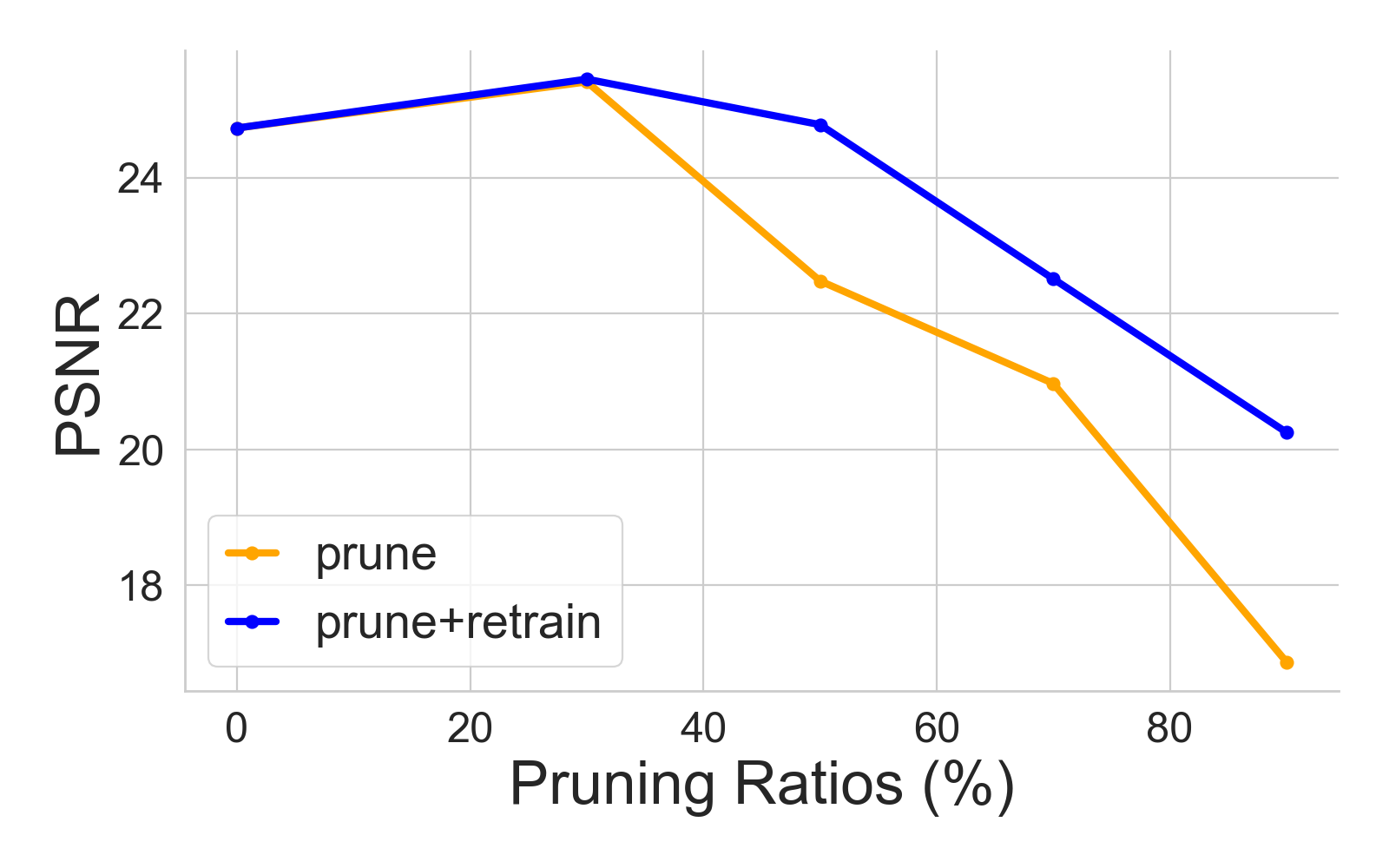}}
        \subfigure[MSE for the fern dataset. ]{\includegraphics[width=.45\textwidth]{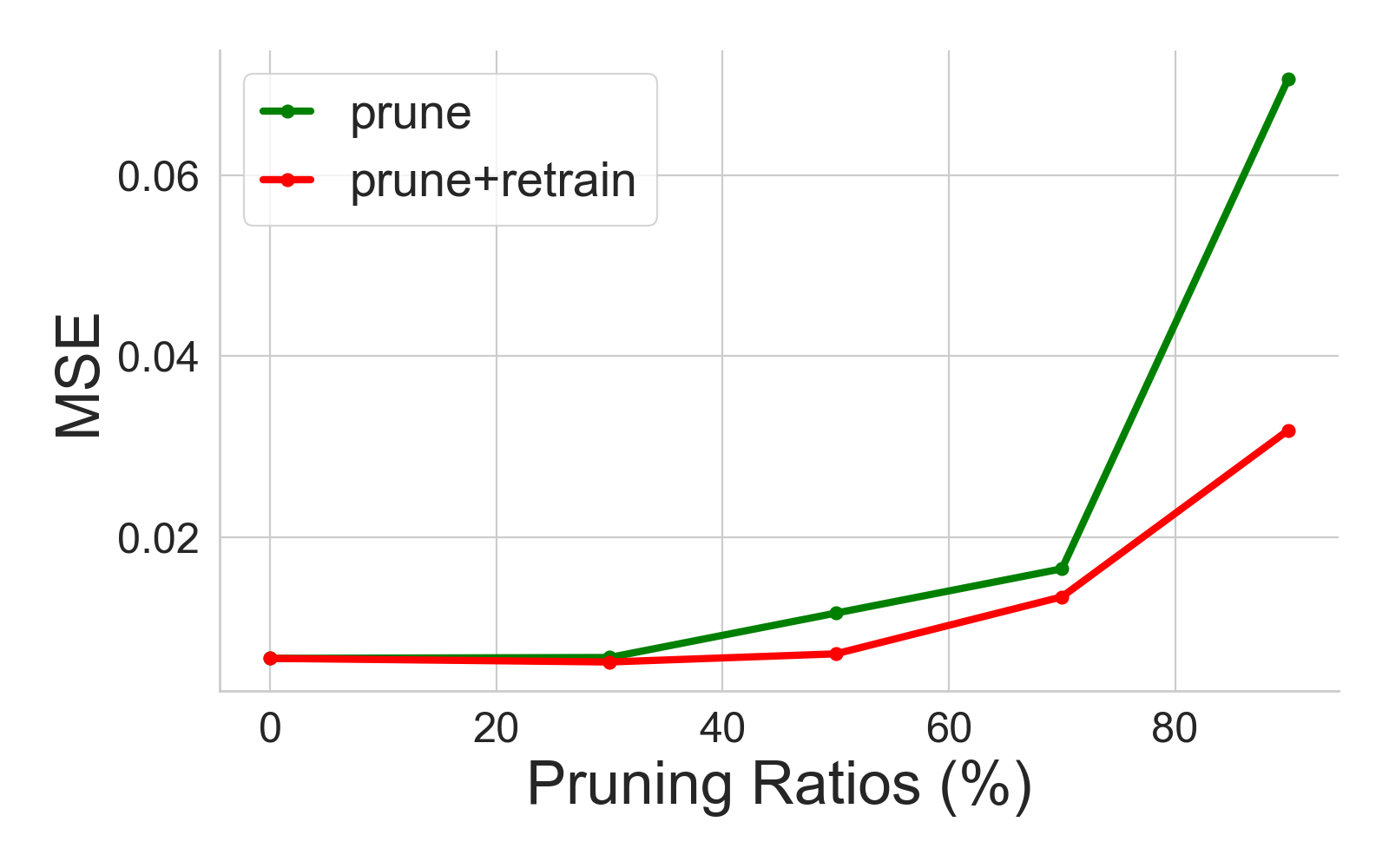}}
   \caption{PSNR and MSE for the fern dataset.}
\label{fig:psnr_mse_fern}
\end{figure}

\begin{figure}[!h]
        \centering
        \subfigure[PSNR for the lego dataset. ]{\includegraphics[width=.45\textwidth]{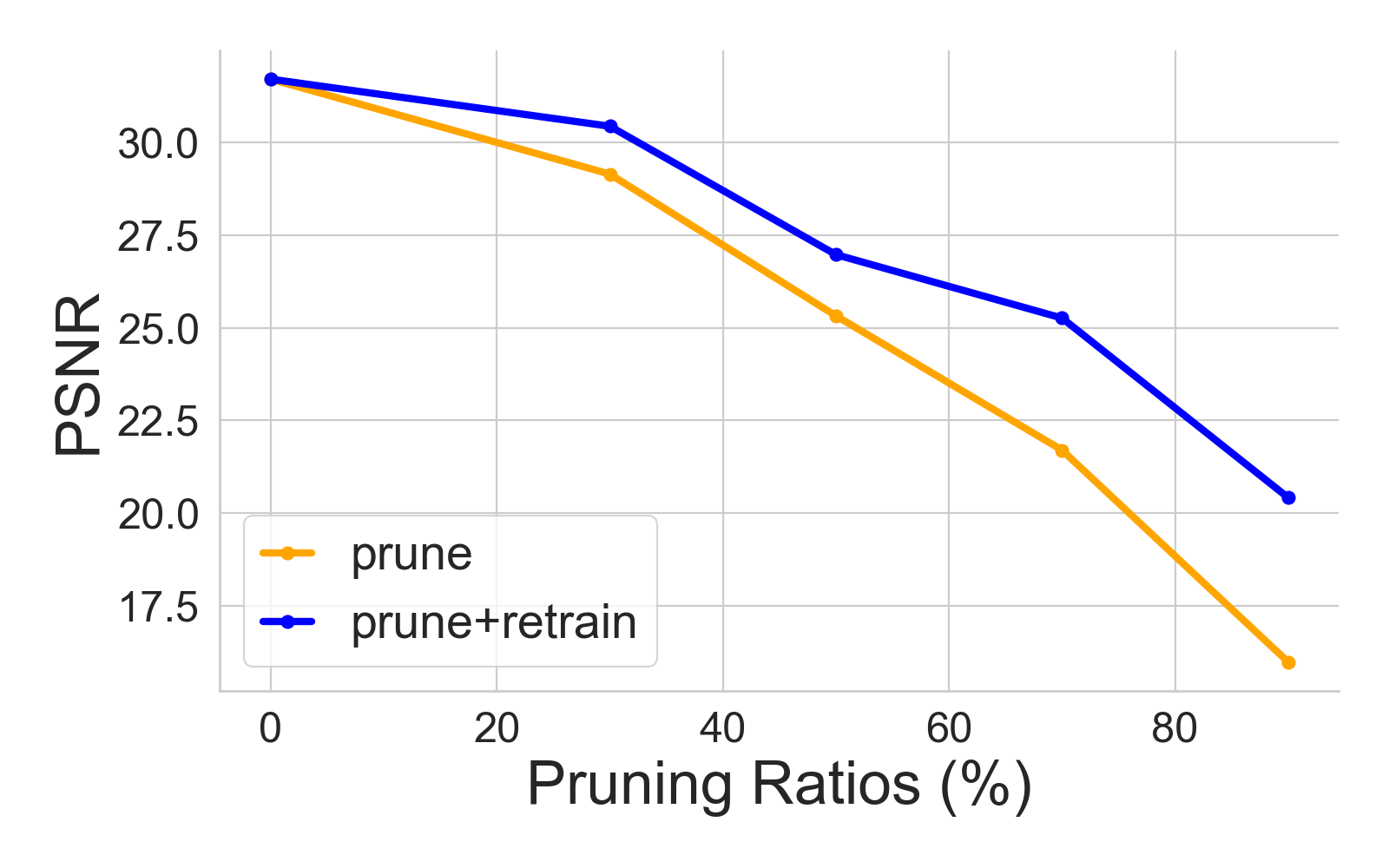}}
        \subfigure[MSE for the ego dataset. ]{\includegraphics[width=.45\textwidth]{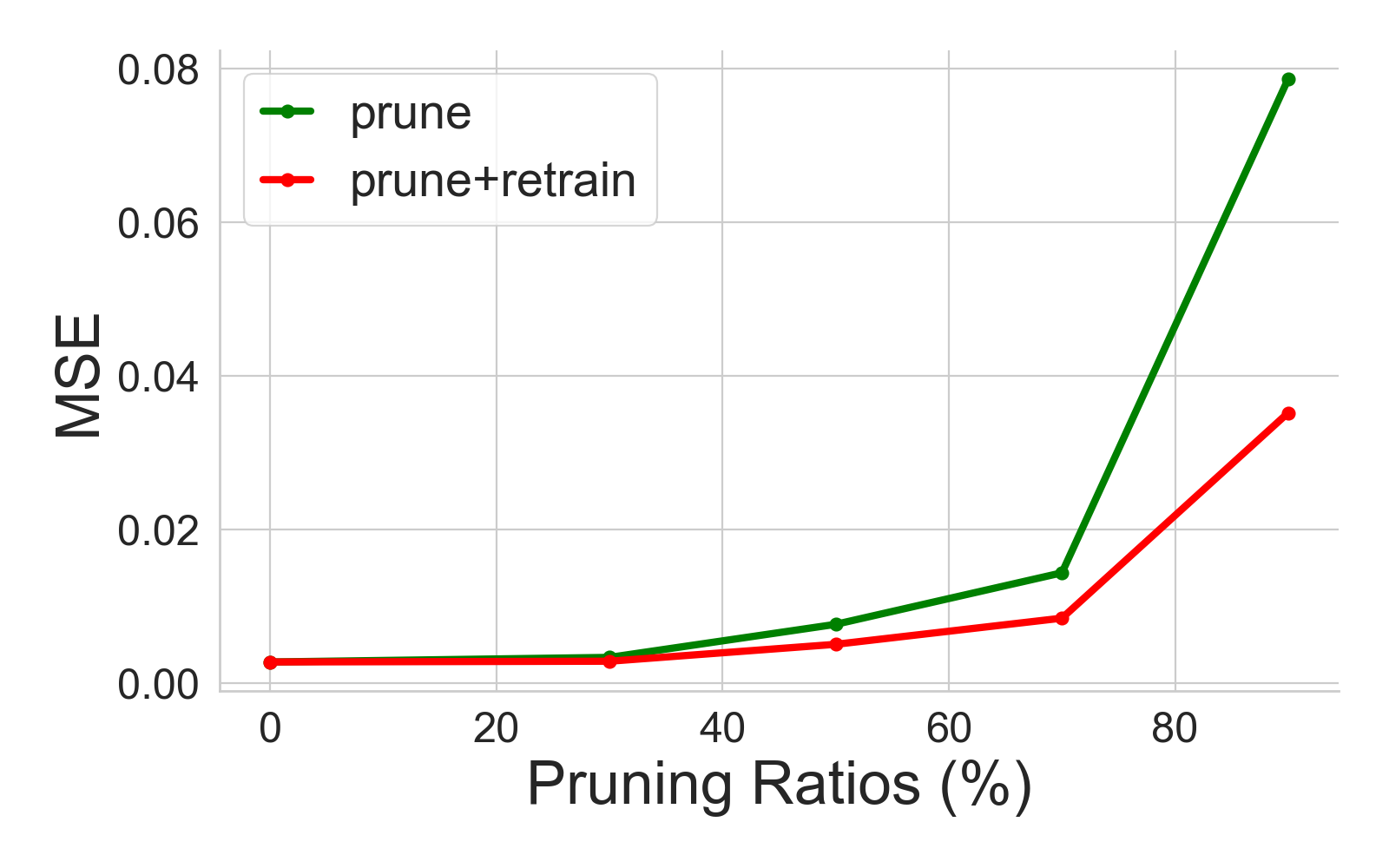}}
   \caption{PSNR and MSE for the lego dataset}
\label{fig:psnr_mse_lego}
\end{figure}

\begin{figure*}[!t]
        \centering
        \subfigure[$30\%$ Pruning. ]{\includegraphics[width=.35\textwidth]{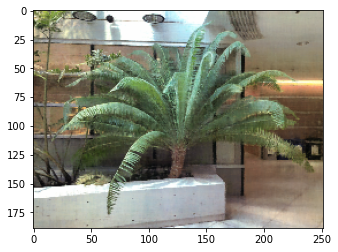}}
        \subfigure[$30\%$ Pruning+Retraining. ]{\includegraphics[width=.35\textwidth]{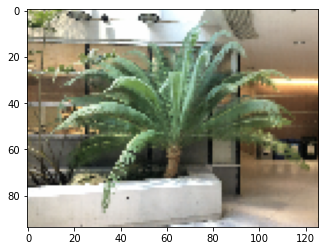}}
        \subfigure[$50\%$ Pruning. ]{\includegraphics[width=.35\textwidth]{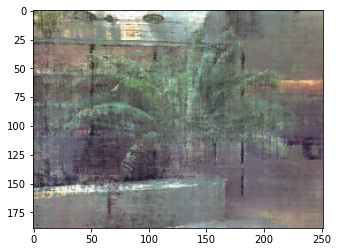}}
        \subfigure[$50\%$ Pruning+Retraining. ]{\includegraphics[width=.35\textwidth]{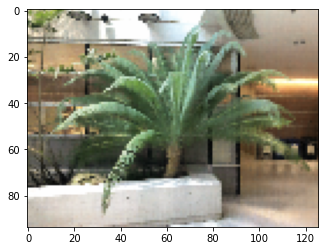}}
        \subfigure[$70\%$ Pruning. ]{\includegraphics[width=.35\textwidth]{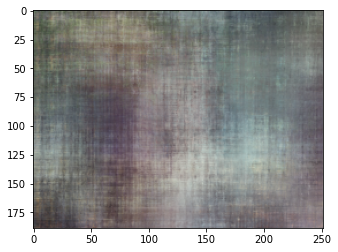}}
        \subfigure[$70\%$ Pruning+Retraining. ]{\includegraphics[width=.35\textwidth]{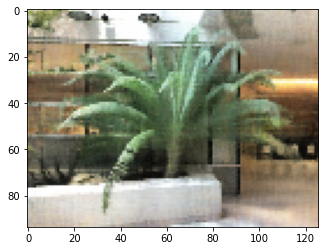}}
        \subfigure[$90\%$ Pruning. ]{\includegraphics[width=.35\textwidth]{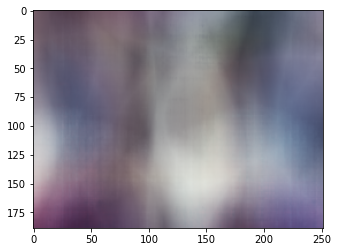}}
        \subfigure[$90\%$ Pruning+Retraining. ]{\includegraphics[width=.35\textwidth]{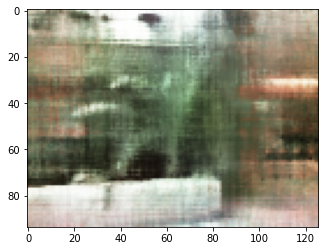}}
        
   \caption{Rendered views of the fern dataset using pruned NeRF models. (a), (c), (e), (g): without retraining after pruning; (b), (d), (f), (h): with retraining after pruning.}
\label{fig:fern_colored}
\end{figure*}

\begin{figure*}[!t]
        \centering
        \subfigure[$30\%$ Pruning. ]{\includegraphics[width=.35\textwidth]{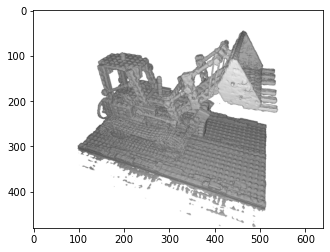}}
        \subfigure[$30\%$ Pruning+Retraining. ]{\includegraphics[width=.35\textwidth]{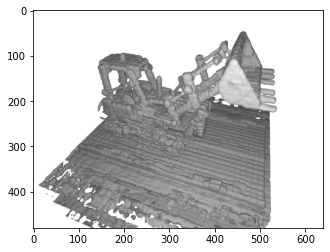}}
        \subfigure[$50\%$ Pruning. ]{\includegraphics[width=.35\textwidth]{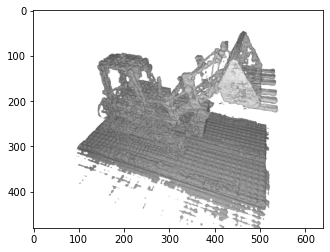}}
        \subfigure[$50\%$ Pruning+Retraining. ]{\includegraphics[width=.35\textwidth]{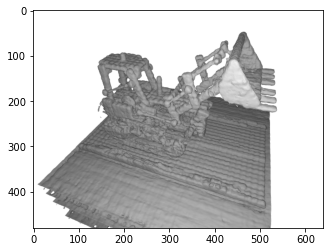}}
        \subfigure[$70\%$ Pruning. ]{\includegraphics[width=.35\textwidth]{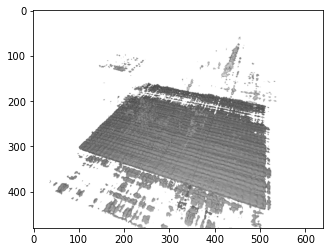}}
        \subfigure[$70\%$ Pruning+Retraining. ]{\includegraphics[width=.35\textwidth]{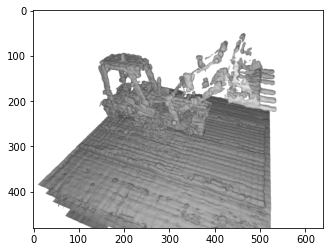}}
        \subfigure[$90\%$ Pruning. ]{\includegraphics[width=.35\textwidth]{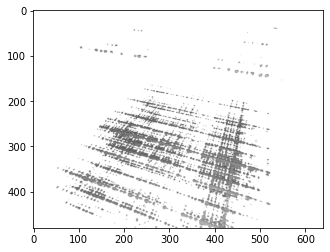}}
        \subfigure[$90\%$ Pruning+Retraining. ]{\includegraphics[width=.35\textwidth]{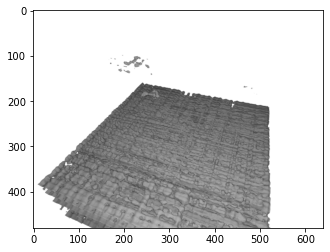}}
   \caption{Mesh models of the lego dataset rendered from pruned NeRF models.(a), (c), (e), (g): without retraining after pruning; (b), (d), (f), (h): with retraining after pruning.}
\label{fig:lego_mesh}
\end{figure*}

\begin{figure*}[!t]
        \centering
        \subfigure[$30\%$ Pruning. ]{\includegraphics[width=.35\textwidth]{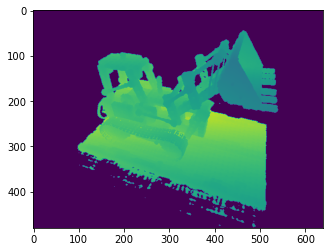}}
        \subfigure[$30\%$ Pruning+Retraining. ]{\includegraphics[width=.35\textwidth]{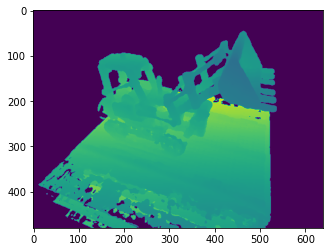}}
        \subfigure[$50\%$ Pruning. ]{\includegraphics[width=.35\textwidth]{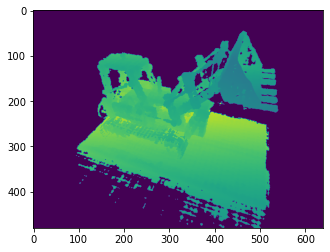}}
        \subfigure[$50\%$ Pruning+Retraining. ]{\includegraphics[width=.35\textwidth]{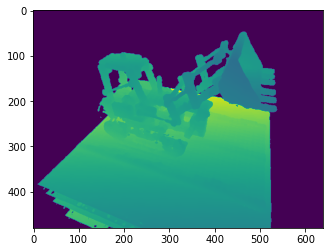}}
        \subfigure[$70\%$ Pruning. ]{\includegraphics[width=.35\textwidth]{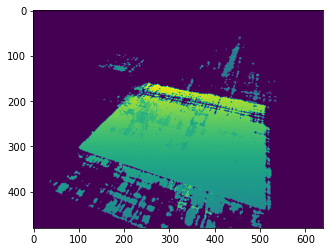}}
        \subfigure[$70\%$ Pruning+Retraining. ]{\includegraphics[width=.35\textwidth]{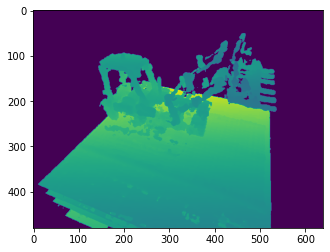}}
        \subfigure[$90\%$ Pruning. ]{\includegraphics[width=.35\textwidth]{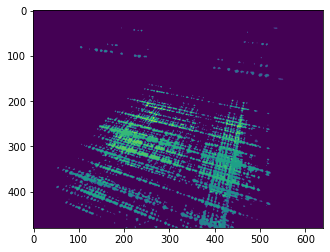}}
        \subfigure[$90\%$ Pruning+Retraining. ]{\includegraphics[width=.35\textwidth]{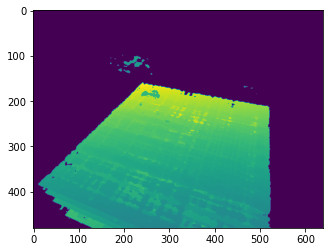}}
   \caption{Depth maps of the lego dataset rendered from pruned NeRF models.(a), (c), (e), (g): without retraining after pruning; (b), (d), (f), (h): with retraining after pruning.}
\label{fig:lego_depth}
\end{figure*}

We give the original 3D reconstructions in Figure~\ref{fig:originals} where we present a rendered view from the fern dataset, together with a mesh model and a depth map of the lego dataset. Notice that the background in the fern dataset is not very suitable for mesh models and depth maps. That is why we do not present mesh models and depth maps of the fern dataset. Figure~\ref{fig:fern_colored} shows the colored views of the fern dataset rendered from the pruned and retrained NeRF models. It is seen that $30\%$ pruning does not produce significant artifacts. However, after $50\%$ pruning, both the geometry and coloring have easily visible artifacts. The quality degradation is even more disturbing after $70\%$ pruning. With the retraining step followed by pruning, these artifacts are mostly avoided except for $90\%$ pruning. \\

Figures~\ref{fig:lego_mesh}-\ref{fig:lego_depth} show the meshes and depth maps rendered from pruned and retrained NeRF models for lego dataset. As can be seen from the figures, the $30\%$ pruned NeRF model is able to reconstruct almost the same mesh and depth map as the original model visually. While we can detect only a few artifacts on the mesh and depths rendered by the $50\%$ pruned NeRF model, the $70\%$ pruned model produces a mesh model and a depth map, which are quite different from the original reconstructions in Figure~\ref{fig:originals} with many missing parts. We see that retraining increases the quality of both mesh models and depth maps by filling the missing parts. One interesting observation is, compared to the original mesh model and depth map in Figure~\ref{fig:originals}, retraining provides a denser reconstruction by filling the empty spaces in the original reconstructions. \\

\setlength{\tabcolsep}{3pt}
\begin{table}[h]
\centering
\resizebox{\columnwidth}{!}{ 
\begin{tabular}{ccccc}
\toprule
Prun. Ratio & Model & PSNR & MSE & Compression\\
\midrule
$0 \% $ & Original &  $24.73$& $0.0066$ & $\times 1.00$\\
\midrule
\centered{$30\%$} & \centered{Pruned\\
Retrained} & \centered{$25.41$ \\ $25.45$} & \centered{$0.0067$ \\ $0.0062$} & \centered{$\times 1.43$} \\
\midrule
\centered{$50\%$}  & \centered{Pruned \\ Retrained} & \centered{$22.48$\\$24.78$} & \centered{$0.0116$\\$0.0071$} & \centered{$\times 2.00$}\\
\midrule
\centered{$70\%$} & \centered{Pruned \\ Retrained} & \centered{$20.97$\\$22.51$} & \centered{$0.0165$\\$0.0134$} & \centered{$\times 3.33$}\\
\midrule
\centered{$90\%$} & \centered{Pruned \\ Retrained} & \centered{$16.87$ \\ $20.25$} & \centered{$0.0706$ \\ $0.0318$} & \centered{$\times 10.00$}\\
\bottomrule
\\
\end{tabular}
}
\caption{Quantitative comparison of pruned models with the original model (fern dataset).
}
\label{tab:comparison_fern}
\end{table}

\setlength{\tabcolsep}{3pt}
\begin{table}[h]
\centering
\resizebox{\columnwidth}{!}{
\begin{tabular}{ccccc}
\toprule
Prun. Ratio & Model & PSNR & MSE & Compression \\
\midrule
$0\%$ & Original & 31.70 & 0.0028 & $\times 1.00$\\
\midrule
\centering{$30\%$} & \centered{Pruned \\ Retrained} & \centered{29.14\\30.43} & \centered{0.0034\\0.0029} & \centered{$\times 1.43$}\\
\midrule
\centered{$50\%$} & \centered{Pruned \\ Retrained} & \centered{25.32\\ 26.97} & \centered{0.0077 \\ 0.0051} & \centered{$\times 2.00$}\\
\midrule 
\centered{$70\%$} & \centered{Pruned\\ Retrained} & \centered{21.70\\ 25.26} & \centered{0.0144\\0.0085} & \centered{$\times 3.33$} \\
\midrule 
\centered{$90\%$} & \centered{Pruned\\ Retrained} & \centered{15.99\\ 20.42} & \centered{0.0786 \\ 0.0352} & \centered{$\times 10.00$} \\
\bottomrule
\\
\end{tabular}
}
\caption{Quantitative comparison of pruned models with the original model (lego dataset).
}
\label{tab:comparison_lego}
\end{table}

In Tables~\ref{tab:comparison_fern}-\ref{tab:comparison_lego}, we give the quantitative evaluation of the proposed compression method. As expected, PSNR decreases, and MSE increases as the pruning ratio increases with one exception at $30\%$ pruning on the fern dataset in Table~\ref{tab:comparison_fern}. This behavior is common in almost every neural network compression technique: at some moderate compression ratio, we see an improvement in the performance. This can be explained by the denoising nature of compression. We also present the plots of these numbers in Figures~\ref{fig:psnr_mse_fern}-\ref{fig:psnr_mse_lego}. \\

Finally, we share videos showing different views of the reconstructed 3D scene (mesh models for the lego dataset and views from the fern dataset), rendered by the original and pruned NeRF models, in  \url{https://drive.google.com/drive/folders/1DmkVXWOLdcDOiD-ed1RHqxACEWNM5uMH?usp=sharing}.

\section{Conclusion}
We propose a straightforward yet novel 3D compression method based on compressing a neural functional representation of 3D scenes. In our experiments, we have observed that there is no considerable quality degradation in the renderings with $30\%$ pruning. However, as the pruning ratio increases to $90\%$, the reconstructions become poorer with visible artifacts. Fortunately, retraining after pruning fixes most artifacts and improves the quality of the reconstructions.  

\clearpage
{\small
\bibliographystyle{ieee_fullname}
\bibliography{cvpr}
}

\end{document}